\documentclass{article}

\usepackage{times}
\usepackage{graphicx} 
\usepackage{subfigure}

\usepackage{natbib}

\usepackage{algorithm}
\usepackage{algorithmic}

\usepackage{hyperref}

\usepackage[accepted]{icml2016}

\newcommand{\mytitle}{Model-Free Imitation Learning with Policy Optimization}
\icmltitlerunning{\mytitle}

\usepackage{amsmath,amssymb,amsthm,mathtools}

\newcommand\bbP{\ensuremath{\mathbb{P}}} 
\newcommand\bbE{\ensuremath{\mathbb{E}}} 
\DeclareMathOperator*{\minimize}{minimize}
\newcommand{\suchthat}{\;\ifnum\currentgrouptype=16 \middle\fi|\;}

\newcommand\bbA{\ensuremath{\mathbb{A}}} 
\newcommand{\pie}{{\pi_E}}
\newcommand{\piold}{{\pi_0}} 
\newcommand{\thetaold}{{\theta_0}} 
\newcommand{\ctrue}{{c_{\text{true}}}}

\newcommand{\Mold}{M}
\newcommand{\Lold}{L}
\newcommand{\psiold}{\psi}

\DeclarePairedDelimiterX{\infdivx}[2]{(}{)}{%
  #1\;\delimsize\|\;#2%
}
\newcommand{\kl}{D_\mathrm{KL}\infdivx}
\newcommand{\klbar}{\overline{D}_{\mathrm{KL}}\infdivx}

\newcommand{\jkg}[1]{}
\begin{document}

\twocolumn[
\icmltitle{\mytitle}

\icmlauthor{Jonathan Ho}{hoj@cs.stanford.edu}
\icmlauthor{Jayesh K. Gupta}{jkg@cs.stanford.edu}
\icmlauthor{Stefano Ermon}{ermon@cs.stanford.edu}
\icmladdress{Stanford University}

\icmlkeywords{imitation learning, apprenticeship learning, inverse reinforcement learning, inverse optimal control}

\vskip 0.3in
]

\begin{abstract}
 In imitation learning, an agent learns how to behave in an environment with an unknown cost function by mimicking expert demonstrations.
 Existing imitation learning algorithms typically involve solving a sequence of planning or reinforcement learning problems. Such algorithms are therefore not directly applicable to large, high-dimensional environments, and their performance can significantly degrade if the planning problems are not solved to optimality.
Under the apprenticeship learning formalism, we develop alternative model-free algorithms for finding a parameterized stochastic policy that performs at least as well as an expert policy on an unknown cost function, based on sample trajectories from the expert. Our approach, based on policy gradients, scales to large continuous environments with guaranteed convergence to local minima.
\end{abstract}

\section{Introduction}


To use reinforcement learning, the learner needs access to a cost or reward signal to identify desirable outcomes. 
The dependence between the cost function and the corresponding optimal policy, however, is generally complex, as it involves planning. In practice, eliciting a cost function that achieves desired behavior can be difficult~\cite{bagnell2015invitation}. An alternative and often more practical approach, called \emph{imitation learning}, is to encode preferences and differentiate between desirable and undesirable outcomes using demonstrations provided by an expert~\cite{pomerleau1991efficient,russell1998learning}.

The simplest approach to imitation learning is behavioral cloning, in which the goal is to learn the relationship between states and optimal actions as a supervised learning problem~\cite{pomerleau1991efficient}. While conceptually simple and theoretically sound~\cite{syed2010reduction}, small inaccuracies of the learned model compound over time, and can lead to situations that are quite different from the ones encountered during training. This is often referred to as the problem of \emph{cascading errors}, and is related to \emph{covariate shift}~\cite{ross2010efficient,bagnell2015invitation}.

Inverse reinforcement learning (IRL) methods \cite{russell1998learning,ng2000algorithms,Ratliff2006-bm,ziebart2008maximum}, which are some of the most successful approaches to imitation learning, assume that the behavior the learner desires to imitate is generated by an expert behaving optimally with respect to an unknown cost function.
IRL algorithms train models over entire trajectories of behavior instead of individual actions, and hence do not suffer from cascading error problems. Furthermore, because the assumption of expert optimality acts as a prior on the space of policies, IRL algorithms can allow the learner to generalize expert behavior to unseen states much more effectively than if the learner had tried to produce a policy or value function instead~\cite{ng2000algorithms,bagnell2015invitation}. 

Unfortunately, the assumption of expert optimality leads to expensive design choices in IRL algorithms. At each iteration, to determine whether a certain cost function $c$ fits an expert policy $\pie$, the IRL algorithm must compare the return of $\pie$ with the return of all other possible policies. Most IRL algorithms do this by running a reinforcement learning algorithm on $c$~\cite{Neu2009-xe}. Because reinforcement learning must be run at each iteration, IRL algorithms can be extremely expensive to run in large domains.

We forgo learning a cost function. We propose a method that directly learns a policy from expert trajectories,
exploiting for learning signal a \emph{class} of cost functions, which
distinguish the expert policy from all others. We first develop a simple,
unified view of a certain class of imitation learning algorithms called
\emph{apprenticeship learning} algorithms~\citep{Abbeel2004-cf,syed2008apprenticeship}. This view naturally leads to the
development of a gradient-based optimization formulation over parameterized
policies for apprenticeship learning. We then provide two model-free
realizations of these optimization algorithms: one is based on a standard policy
gradient algorithm, and the other is based on a recently developed policy gradient algorithm that incorporates trust region constraints to stabilize optimization. We demonstrate the effectiveness of our approach on control problems with very high-dimensional observations (over 600 continuous features), for which we train neural network control policies from scratch.

\section{Preliminaries}

We begin by defining basic notions from reinforcement learning. We are given an environment consisting of a state space $\mathcal{S}$, an action space $\mathcal{A}$, a dynamics model $p(s'|s,a)$, and an initial state distribution $p_0(s_0)$. Agents act according to stationary stochastic policies $\pi(a|s)$, which specify action choice probabilities for each state. We will work with finite $\mathcal{S}$ and $\mathcal{A}$, but our methods will extend to continuous spaces.

With respect to a cost function $c:\mathcal{S}\times\mathcal{A} \rightarrow \mathbb{R}$, a discount factor $\gamma \in [0,1)$, and a policy $\pi$, the \emph{state-action value} function $Q_\pi^c(s,a)$, the \emph{state value} function $V_\pi^c(s)$, and the \emph{advantage} function $A_\pi^c(s,a)$ are defined as ${Q_\pi^c(s_t,a_t) = \bbE_{p_0,p,\pi}\left[\sum_{t'=t}^\infty \gamma^{t'-t} c(s_{t'},a_{t'})\right]}$,
$V_\pi^c(s) = \bbE_{a\sim\pi(\cdot|s)}\left[Q_\pi^c(s,a)\right]$, and
$A_\pi^c(s,a) = Q_\pi^c(s,a) - V_\pi^c(s)$.
The \emph{expected cost} of $\pi$ is $\eta^c(\pi) = \bbE_{s_0\sim p_0}\left[V_\pi^c(s_0)\right]$.
For clarity, when $c$ is a parameterized function, written as $c_w$ for a parameter vector $w$, we will replace $c$ by $w$ in the names of these quantities---for example, $\eta^w$, $Q_\pi^w$, etc. 
We define
the $\gamma$-discounted \emph{state visitation distribution} of a policy $\pi$ by $\rho_\pi(s) = \sum_{t=0}^\infty \gamma^t  \bbP_{p_0, \pi}\left[ s_t = s \right]$,
where $\bbP_{p_0, \pi}\left[ s_t = s \right]$ is the probability of landing in state $s$ at time $t$, when following $\pi$ starting from $s_0 \sim p_0$. When convenient, we will overload notation for \emph{state-action visitation distributions}: $\rho_\pi(s,a) = \pi(a|s)\rho_\pi(s)$, allowing us to write expected cost as $\eta^c(\pi) = \sum_{s,a} \rho_\pi(s,a)\,c(s,a) = \bbE_{\rho_\pi}[c(s,a)]$.

\section{Apprenticeship learning}
\label{sec:apprenticeshiplearning}

To address the imitation learning problem, we adopt the apprenticeship learning formalism, in which the learner must find a policy that performs at least as well as the expert $\pie$ on an unknown true cost function ${\ctrue:\mathcal{S} \times \mathcal{A} \rightarrow \mathbb{R}}$~\cite{Abbeel2004-cf,syed2008apprenticeship}. Formally, the learner's goal is to find a policy $\pi$ such that $\eta^{\ctrue}(\pi) \leq \eta^{\ctrue}(\pie)$, given a dataset of trajectory samples from $\pie$.

To do this, apprenticeship learning algorithms carry with them the assumption that the true cost function belongs to a class of cost functions $\mathcal{C}$. Accordingly, they seek a policy $\pi$ that performs as well as $\pie$ for all $c \in \mathcal{C}$---that is, they seek to satisfy the constraints
\begin{align}
  \eta^{c}(\pi) \leq \eta^{c}(\pie) \quad\text{for all}\quad c \in \mathcal{C}\label{eq:alconstraints}
\end{align}
Because $\ctrue \in \mathcal{C}$ by assumption, satisfying this family of constraints ensures successful apprenticeship learning. We can reformulate this constraint satisfaction problem as an optimization problem by defining the objective
\begin{align}
\delta_\mathcal{C}(\pi, \pie) = \sup_{c \in \mathcal{C}} \eta^{c}(\pi) - \eta^{c}(\pie) \label{eq:alobjective}
\end{align}
Intuitively, the cost functions in $\mathcal{C}$ distinguish the expert from all other policies, assigning high expected cost to non-expert policies and low expected cost to the expert policy. If $\delta_c(\pi,\pie) > 0$, then there exists some cost in $\mathcal{C}$ such that $\pi$ performs worse than $\pie$---in this case, $\pi$ is a poor solution to the apprenticeship learning problem. On the other hand, if $\delta_c(\pi,\pie) \leq 0$, then $\pi$ performs at least as well as $\pie$ for all costs in $\mathcal{C}$, and therefore satisfies the apprenticeship learning constraints~\eqref{eq:alconstraints}.


Having defined the objective, the job of an apprenticeship learning algorithm is to solve the optimization problem
\begin{align}
  \minimize_\pi\ \delta_\mathcal{C}(\pi, \pie). \label{eq:alprob}
\end{align}
So far, we have described a general framework for defining apprenticeship
learning algorithms. To instantiate this framework, two ingredients must be
provided: a cost function class $\mathcal{C}$, and an optimization algorithm to
solve~\eqref{eq:alprob}. Our goal in this paper is to address the optimization
ingredient, so we will use linearly parameterized cost functions in our experiments, although the development of our method is agnostic to the particulars of the cost function class.
In Section~\ref{sec:policyopt}, we will develop a method for approximately solving~\eqref{eq:alprob} over a class of parameterized stochastic policies (for example, neural network policies), assuming generic access to a method for solving the maximization~\eqref{eq:alobjective} over costs for fixed policies $\pi$. Our method will perform gradient-based stochastic optimization on policy parameters---we refer to this strategy as \emph{policy optimization}. 

\subsection{Examples from prior work}

Before delving into our method, we first review two prototypical examples of apprenticeship learning algorithms. We show how they fall into the framework detailed in this section (namely, how they choose the cost function class $\mathcal{C}$), and we briefly describe their solution techniques for solving~\eqref{eq:alprob}, which differ vastly from our new policy optimization method.

\paragraph{Feature expectation matching}
\label{sec:featexpmatch}

\citet{Abbeel2004-cf} define $\mathcal{C}$ by first fixing a set of basis cost functions $c_1, \dotsc, c_k$, where $c_j:\mathcal{S}\times \mathcal{A} \rightarrow \mathbb{R}$, and then defining the cost class as a certain set of linear combinations of these basis functions:
\begin{align}\textstyle
  \mathcal{C}_\text{linear} = \left\{ c_w \triangleq \sum_{i=1}^k w_i c_i \suchthat \|w\|_2 \leq 1 \right\}\label{eq:linearcost}
\end{align}
The structure of $\mathcal{C}_\text{linear}$ allows the expected costs with respect to $c_w$ to be written as an inner product of $w$ with a certain \emph{feature expectation} vector of $\pi$, defined as $\phi(\pi) \triangleq \bbE_{\rho_\pi}\left[ \sum_{t=0}^\infty \gamma^t \phi(s_t,a_t) \right]$, where $\phi(s,a) = [c_1(s,a) \cdots c_k(s,a)]^T$. Because any cost in $\mathcal{C}_\text{linear}$ can be written as $c_w(s,a) = w\cdot\phi(s,a)$, linearity of expectation yields $
  \eta^w(\pi) = \bbE\left[ \sum_{t=0}^\infty \gamma^t w \cdot \phi(s_t,a_t) \right] = w \cdot \phi(\pi)$.

Based on this observation, \citet{Abbeel2004-cf} propose to match feature expectations; that is, to find a policy $\pi$ such that $\phi(\pi) \approx \phi(\pie)$, thereby guaranteeing that $\eta^w(\pi) = w\cdot\phi(\pi) \approx w\cdot\phi(\pie) = \eta^w(\pie)$ for all cost functions $c_w \in \mathcal{C}_\text{linear}$.
We can understand feature expectation matching as minimization of $\delta_{\mathcal{C}_\text{linear}}$, because
\begin{align}
\delta_{\mathcal{C}_\text{linear}}(\pi, \pie) &= \!\!\sup_{\|w\|_2 \leq 1} \bbE_{\rho_\pi}[w\cdot\phi(s,a)] - \bbE_{\rho_{\pie}}[w\cdot\phi(s,a)] \nonumber \\
&= \!\!\sup_{\|w\|_2 \leq 1} w \cdot (\phi(\pi) - \phi(\pie)) \nonumber \\
&= \|\phi(\pi) - \phi(\pie)\|_2. \label{eq:matchfeatexp}
\end{align}

To solve this problem, \citet{Abbeel2004-cf} propose to incrementally generate a
set of policies by inverse reinforcement learning~\cite{ng2000algorithms}. At
each iteration, their algorithm finds a cost function that assigns low expected
cost to the expert and high expected cost to previously found policies. Then, it
adds to the set of policies the optimal policy for this cost, computed via
reinforcement learning. These steps are repeated until there is no cost function
that makes the expert perform much better than the previously found policies. The final policy, which minimizes \eqref{eq:matchfeatexp}, is produced by stochastically mixing the generated policies using weights calculated by a quadratic program. This algorithm is quite expensive to run for large MDPs, because it requires running reinforcement learning at each iteration.

\paragraph{Game-theoretic approaches}\label{sec:gametheoretic}
\citet{syed2007game,syed2008apprenticeship} proposed two apprenticeship learning algorithms, Multiplicative Weights Apprenticeship Learning (MWAL) and Linear Programming Apprenticeship Learning (LPAL), that also use basis cost functions, but with the weights restricted to give a convex combination:
\begin{align}\textstyle
  \mathcal{C}_\text{convex} = \left\{ c_w \triangleq \sum_{i=1}^k w_i c_i \suchthat w_i \geq 0, \ \sum_i w_i = 1 \right\}
\end{align}
MWAL uses a multiplicative weights update method to solve the resulting optimization problem, and like Abbeel and Ng's method, requires running reinforcement learning in its inner loop. \citet{syed2008apprenticeship} address this computational complexity with their LPAL method. They notice that restricting the weights on the basis functions to lie on the simplex allows the maximization over costs to be performed instead over a finite set: the problem~\eqref{eq:alprob} with $\mathcal{C}_\text{convex}$ can be written as $\min_\pi \max_{i\in\{1,\dotsc,k\}} \eta^{c_i}(\pi) - \eta^{c_i}(\pie)$.
Syed et al. are therefore able to formulate a single linear program on state-action visitation frequencies that simultaneously encodes~\eqref{eq:alprob} and Bellman flow constraints on the frequencies to ensure that they can be generated by some policy in the environment.

Inspired by LPAL, we will formulate an unconstrained optimization approach to apprenticeship learning. We propose to optimize~\eqref{eq:alprob} directly over parameterized stochastic policies instead of state-action visitation distributions, allowing us to scale to large spaces without keeping variables for each state and action. We keep our formulation general enough to allow for any cost function class $\mathcal{C}$, but because the focus of this paper is the optimization over policies, we will use linearly parameterized cost classes $\mathcal{C}_\text{linear}$ and $\mathcal{C}_\text{convex}$ in our experiments. (Note that despite assuming linearity, this setting is already more general than LPAL, as the maximization over $\mathcal{C}_\text{linear}$ cannot be written as a maximization over a finite set.)

\section{Policy optimization for apprenticeship learning}
\label{sec:policyopt}
Having reviewed existing algorithms for solving various settings of apprenticeship learning, we now delve into policy optimization strategies that directly operate on~\eqref{eq:alprob} as a stochastic optimization problem. To allow us to scale to large, continuous environments, we first fix a class of smoothly parameterized stochastic policies $\Pi = \{\pi_\theta \suchthat \theta\in \Theta \}$, where $\Theta$ is a set of valid parameter vectors. With this class of policies, our goal is to solve the optimization problem~\eqref{eq:alprob} over policy parameters:
\[ \minimize_\theta\ \delta_\mathcal{C}(\pi_\theta, \pie) \]
where $\delta_\mathcal{C}(\pi_\theta, \pie) = \sup_{c \in \mathcal{C}} \eta^{c}(\pi_\theta) - \eta^{c}(\pie)$.
We propose find a local minimum to this problem using gradient-based stochastic optimization. To this end, let us first examine the gradient of $\delta_\mathcal{C}$ with respect to $\theta$. Letting $c^*$ denote the cost function that achieves the supremum in $\delta_C$,\footnote{In this paper, we only work with classes $\mathcal{C}$ for which this supremum is achieved.} we have
\begin{align}
\nabla_\theta \delta_\mathcal{C}(\pi_\theta, \pie) = \nabla_\theta \eta^{c^*}(\pi_\theta) \label{eq:algrad}
\end{align}
This formula dictates the basic structure a gradient-based algorithm must take to minimize $\delta_\mathcal{C}$---it must first compute $c^*$ for a fixed $\theta$, then it must use this $c^*$ to improve $\theta$ for the next iteration. (Our algorithm in Section~\ref{sec:trpoal} will actually have to identify a cost function defined by a more complicated criterion, but as we will see in Section~\ref{sec:cost}, this will not pose significant difficulty.) Because the cost $c^*$ effectively defines a reinforcement learning problem at the current policy $\pi_\theta$, we can interpret gradient-based optimization of $\delta_\mathcal{C}$ as a procedure that alternates between (1) fitting a local reinforcement learning problem to generate learning signal for imitation, and (2) improving the policy with respect to this local problem. We will discuss how to find $c^*$ in Section~\ref{sec:cost}; for now, we will only discuss strategies for policy improvement.

\subsection{Policy gradient}
\label{sec:policygrad}

The most straightforward method of optimizing~\eqref{eq:alprob} is stochastic gradient descent with an estimate of the gradient~\eqref{eq:algrad}:
\begin{align}
\nabla_\theta \eta^{c^*}(\pi_\theta) = \bbE_{\rho_{\pi_\theta}} \left[ \nabla_\theta \log\pi_\theta(a|s) Q_{\pi_\theta}^{c^*}(s,a) \right] \label{eq:reinforce}
\end{align}
This is the classic policy gradient formula for the cost function $c^*$~\cite{sutton1999policy}. To estimate this from samples, we propose the following algorithm, called \mbox{IM-REINFORCE}, which parallels the development of REINFORCE~\cite{williams1992simple} for reinforcement learning. As input, IM-REINFORCE is given expert trajectories (that is, rollouts of the expert policy). At each iteration, for the current parameter vector $\thetaold$, trajectories are sampled using $\pi_0 \triangleq \pi_{\theta_0}$. Then, the cost $\hat c$ attaining the supremum of an empirical estimate of $\delta_\mathcal{C}$ is calculated to satisfy:
\begin{align}
  \hat\delta_\mathcal{C}(\piold,\pie) &= \sup_{c\in\mathcal{C}}\hat\bbE_{\rho_{\piold}}[c(s,a)] - \hat\bbE_{\rho_{\pie}}[c(s,a)] \\
  &= \hat\bbE_{\rho_{\piold}}[\hat c(s,a)] - \hat\bbE_{\rho_{\pie}}[\hat c(s,a)] \label{eq:empiricaldelta}
\end{align}
Here, $\hat\bbE$ denotes empirical expectation using rollout samples. We describe how to compute $\hat c$ in detail in Section~\ref{sec:cost}; how to do so depends on $\mathcal{C}$.

With $\hat c$, IM-REINFORCE then estimates the gradient $\nabla_\theta \eta^{\hat c}$ using the formula~\eqref{eq:reinforce}, where the state-action value $Q_{\piold}^{\hat c}(s,a)$ is estimated using discounted future sums of $\hat c$ costs along rollouts for $\piold$. Finally, to complete the iteration, IM-REINFORCE takes a step in the resulting gradient direction, producing new policy parameters $\theta$ ready for the next iteration. These steps are summarized in Algorithm
~\ref{alg:policygrad}.

\setlength{\textfloatsep}{5pt}
\begin{algorithm}[tb]
   \caption{IM-REINFORCE}
   \label{alg:policygrad}
\begin{algorithmic}
   \STATE {\bfseries Input:} Expert trajectories $\tau_E$, initial policy parameters. $\theta_0$
   \FOR{$i = 0, 1, 2, \dotsc$}
   \STATE Roll out trajectories $\tau \sim \pi_{\theta_i}$ 
   \STATE Compute $\hat c$ achieving the supremum in \eqref{eq:empiricaldelta},
   \STATE \ \ \ \ \ with expectations taken over $\tau$ and $\tau_E$ 
   \STATE Estimate the gradient $\nabla_\theta \eta^{\hat c}(\pi_{\theta}) |_{\theta=\theta_i}$ \eqref{eq:reinforce} with $\tau$
   \STATE Use the gradient to take a step from $\theta_i$ to $\theta_{i+1}$
   \ENDFOR
\end{algorithmic}
\end{algorithm}

\subsection{Monotonic policy improvements}
While IM-REINFORCE is straightforward to implement, the gradient estimator~\eqref{eq:reinforce} exhibits extremely high variance, making the algorithm very slow to converge, or even diverge for reasonably large step sizes. This variance issue is not unique to our apprenticeship learning formulation, and is a hallmark difficulty of policy gradient algorithms for reinforcement learning~\cite{peters2008reinforcement}.

The reinforcement learning literature contains a vast number of techniques for
calculating high-quality policy parameter steps based on Monte Carlo estimates
of the gradient. We make no attempt to fully review these techniques here.
Instead, we will directly draw inspiration from a recently developed algorithm called \emph{trust region policy optimization} (TRPO), a model-free policy search algorithm capable of quickly training large neural network stochastic policies for complex tasks~\cite{schulman2015trust}.

\paragraph{TRPO for reinforcement learning}
In this section, we will review TRPO for reinforcement learning, and in the next, we will develop an analogous algorithm for our apprenticeship learning setting. For now, we will drop the cost function superscript $c$, because the cost is fixed in the reinforcement learning setting.

Suppose we have a current policy $\piold$ that we wish to improve. We can write the performance of a new policy $\pi$ in terms of the performance of $\piold$~\cite{kakade2002approximately}:
\begin{align}
  \eta(\pi) = \eta(\piold) + \bbE_{\rho_\pi}\bbE_{a\sim\pi(\cdot|s)} [A_\piold(s,a)]
\end{align}

Vanilla policy gradient methods, such as the one described in the previous section, improve $\eta(\pi)$ by taking a step on a local approximation at $\piold$:
\begin{align}
  \Lold(\pi) \triangleq \eta(\piold) + \bbA_\piold(\pi) 
\end{align}
where $\bbA_\piold(\pi) = \bbE_{s\sim\rho_\piold}\bbE_{a\sim\pi(\cdot|s)} \left[A_\piold(s,a)\right]$.
If the policies are parameterized by $\theta$ (that is $\piold = \pi_{\theta_0}$ and $\pi = \pi_\theta$), then $\Lold$ matches $\eta$ to first order at $\thetaold$, and therefore taking a small gradient step on $\Lold$ guarantees improvement of $\eta$. However, there is little guidance on how large this step can be, and in cases when the gradient can only be estimated, the required step size might be extremely small to compensate for noise. \citet{schulman2015trust} address this by showing that minimizing a certain surrogate loss function can guarantee policy improvement with a large step size. Define the following penalized variant of $\Lold$:
\begin{align}
\Mold(\pi) \triangleq \Lold(\pi) + \frac{2\epsilon\gamma}{(1-\gamma)^2} \max_s\kl{\piold(\cdot|s)}{\pi(\cdot|s)} \label{eq:trpomaj}
\end{align}
where $\epsilon=\max_{s,a} | A_\piold(s,a) |$. \citeauthor{schulman2015trust} prove that  $\Mold$ upper bounds $\eta$:
\begin{align}
\eta(\pi) \leq \Mold(\pi) \label{eq:trpoineq}
\end{align}
Because KL divergence is zero when its arguments are equal, this inequality
shows that $M$ majorizes \footnote{$M$ is said to majorize $\eta$ at $x_0$ if $M \geq \eta$ with equality at $x_0$.} $\eta$ at $\pi_0$. Using $M$ as a majorizer for $\eta$ in a majorization-minimization algorithm leads to an algorithm guaranteeing monotonic policy improvement at each iteration.

Unfortunately, as $M$ is currently defined, computing the maximum-KL divergence term over the whole state space is intractable. \citeauthor{schulman2015trust} propose to relax this to an average over state space, which can be approximated by samples:
\begin{align}
\klbar{\piold}{\pi} \triangleq \bbE_{s\sim\rho_\piold}\left[\kl{\piold(\cdot|s)}{\pi(\cdot|s)}\right] \label{eq:avgkl}
\end{align}
They find that this average-KL formulation works well empirically, and that algorithm's stability could be improved by further reformulating the cost as a trust region constraint. This leads to the TRPO step computation
\begin{align}\begin{split}
  \minimize_\theta \ \Lold(\pi_\theta) \quad \text{s.t.} \quad\klbar{\piold}{\pi_\theta} \leq \Delta
\end{split}
\label{eq:rltrpoprob}
\end{align}
where all constants in Equation~\eqref{eq:trpomaj} are folded into a predefined trust region size $\Delta > 0$.
To solve this step computation problem, the objective $\Lold$ and the KL divergence constraint must be approximated using samples and then minimized with gradient-based constrained optimization. The sample approximation can be done using a similar strategy to the one described in Section~\ref{sec:policygrad}. Further discussion on sampling methodologies and effective optimization algorithms for solving this constrained problem can be found in~\citet{schulman2015trust}.

\paragraph{TRPO for apprenticeship learning}
\label{sec:trpoal}

Now, we describe how to adapt TRPO to apprenticeship learning~\eqref{eq:alprob}. Reintroducing the $c$ superscripts, we wish to compute an improvement step from $\thetaold$ to $\theta$ for the optimization problem
\begin{align}
\minimize_\theta\ \sup_{c \in \mathcal{C}} \eta^{c}(\pi_\theta) - \eta^{c}(\pie)
\end{align}

We wish to derive a majorizer for this objective, analogous to the majorizer~\eqref{eq:trpomaj} for a fixed cost function. Observe that if $\{f_\alpha\}$ and $\{g_\alpha\}$ are families of functions such that $g_\alpha$ majorizes $f_\alpha$ at $x_0$ for all $\alpha$, then $\sup_\alpha g_\alpha$ majorizes $\sup_\alpha f_\alpha$ at $x_0$. We can therefore derive a TRPO-style algorithm for apprenticeship learning as follows. First, to remove the dependence of $\epsilon$ in~\eqref{eq:trpomaj} on any particular cost function, we assume that all cost functions in $\mathcal{C}$ are bounded by $C_\text{max}$,\footnote{In practice, this is easy to satisfy for $\mathcal{C}_\text{linear}$ and $\mathcal{C}_\text{convex}$ by ensuring that the cost basis functions are bounded.} and then we let
$
\epsilon' \triangleq \frac{2C_\text{max}}{1-\gamma} \geq \sup_c\max_{s,a} |A^c_\piold(s,a)|
$. Now, we can define $\Mold^c(\pi)$ analogously to~\eqref{eq:trpomaj}:
\begin{align*}
\Mold^c(\pi) \triangleq \Lold^c(\pi) + \frac{2\epsilon'\gamma}{(1-\gamma)^2} \max_s\kl{\piold(\cdot|s)}{\pi(\cdot|s)}
\end{align*}
By the definition of $\epsilon'$ and~\eqref{eq:trpoineq}, we have that for all $c \in \mathcal{C}$,
\begin{align}
  \eta^c(\pi) - \eta^{c}(\pie) &\leq \Mold^c(\pi) - \eta^{c}(\pie),\label{eq:inequalities}
\end{align}
and consequently, we obtain an upper bound for the apprenticeship objective:
\begin{align}
  \delta_\mathcal{C}(\pi,\pie) &= \sup_{c \in \mathcal{C}} \eta^c(\pi) - \eta^{c}(\pie) \nonumber \\
  &\leq \sup_{c \in \mathcal{C}} \Mold^c(\pi) - \eta^{c}(\pie)\label{eq:supinequality} 
  \triangleq \Mold^\mathcal{C}(\pi, \pie).
\end{align}
Since the inequalities~\eqref{eq:inequalities} become equalities at $\pi = \piold$, inequality~\eqref{eq:supinequality} does too, and thus $\Mold^\mathcal{C}(\pi, \pie)$ majorizes the apprenticeship learning objective $\delta_\mathcal{C}(\pi,\pie)$ at $\pi=\piold$. Importantly, the KL divergence cost in $\Mold^\mathcal{C}(\pi, \pie)$ does not depend on $c$:
\begin{align}
  \Mold^\mathcal{C} (\pi, \pie) = &\left( \sup_{c \in \mathcal{C}} \Lold^c(\pi) - \eta^{c}(\pie) \right) \\
  &+ \frac{2\epsilon'\gamma}{(1-\gamma)^2} \max_s\kl{\piold(\cdot|s)}{\pi(\cdot|s)} \nonumber
\end{align}
Hence, we can apply the same empirically justified transformation that led to TRPO: replacing the maximum-KL cost by an average-KL constraint. We therefore propose to compute steps for our apprenticeship learning setting by solving the following trust region subproblem:
\begin{align}\begin{split}
  \minimize_\theta &\quad\sup_{c \in \mathcal{C}} \Lold^c(\pi_\theta) - \eta^{c}(\pie) \\
  \text{subject to} &\quad\klbar{\piold}{\pi_\theta} \leq \Delta
\end{split}
\label{eq:altrpoproblem}
\end{align}
where again, all constants are folded into $\Delta$.
To solve this trust region subproblem in the finite-sample regime, we approximate the KL constraint using samples from $\piold$, just as TRPO does for its trust region problem~\eqref{eq:rltrpoprob}. The objective of~\eqref{eq:altrpoproblem}, however, warrants more attention, because of the interplay between the maximization over $c$ and minimization over $\theta$. Let $f(\theta)$ be the objective of~\eqref{eq:altrpoproblem}. We wish to derive a finite-sample approximation to $f$ suitable as an objective for the subproblem, so for computational reasons, we would like to avoid trajectory sampling within optimization for this subproblem.

To do so, we introduce importance sampling with $\piold$ as the proposal distribution for the advantage term of $f$, thereby avoiding the need to sample from $\pi_\theta$ as $\theta$ varies:
\begin{align}
f(\theta) &= \sup_{c \in \mathcal{C}} \eta^c(\piold) - \eta^c(\pie) + \bbA^c_{{\piold}}(\pi_\theta)  \nonumber \\
\begin{split}
&= \sup_{c \in \mathcal{C}} \bbE_{\rho_{\piold}}[c(s,a)] - \bbE_{\rho_{\pie}}[c(s,a)]\ + \\
&\qquad\bbE_{\rho_{\piold}}\left[\frac{\pi_\theta(a|s)}{{\piold}(a|s)} (Q^c_{\piold}(s,a) - V^c_{\piold}(s))\right] \label{eq:importance}
\end{split}
\end{align}
At first glance, it seems that the last term of equation~\eqref{eq:importance} requires multiple rollouts for the $Q$ and $V$ parts separately. However, this is not the case, because the identity
\begin{align*}
\bbE_{a\sim\piold(\cdot|s)}\!\!\left[\frac{\pi(a|s)}{\piold(a|s)} V^c_\piold(s) \right]\!
= V^c_\piold(s)
= \bbE_{a\sim\piold(\cdot|s)}[Q^c_\piold(s,a)]
\end{align*}
lets us write
\begin{align}
\begin{split}
f(\theta) &= \sup_{c \in \mathcal{C}} \bbE_{\rho_{\piold}}[c(s,a)] - \bbE_{\rho_{\pie}}[c(s,a)]\ + \\
&\qquad\bbE_{\rho_{\piold}} \left[ \left(\frac{\pi_\theta(a|s)}{{\piold}(a|s)} - 1\right) Q^c_{\piold}(s,a) \right]
\end{split} \label{eq:altrposubprobobj}
\end{align}
Replacing expectations with empirical ones gives the final form of the objective that we use to define the finite-sample trust region subproblem. Solving these trust region subproblems yields our final algorithm, which we call \mbox{IM-TRPO} (Algorithm~\ref{alg:imtrpo}). The computational power needed to minimize this trust region cost is not much greater than that of the TRPO subproblem~\eqref{eq:rltrpoprob}, assuming that the supremum over $\mathcal{C}$ is easily computable. We will show next in Section~\ref{sec:cost} that solving IM-TRPO subproblems indeed poses no significant difficulty over the computation~\eqref{eq:empiricaldelta} necessary for IM-REINFORCE.

\setlength{\textfloatsep}{5pt}
\begin{algorithm}[tb]
   \caption{IM-TRPO}
   \label{alg:imtrpo}
\begin{algorithmic}
   \STATE {\bfseries Input:} Expert trajectories $\tau_E$, initial policy params. $\theta_0$, trust region size $\Delta$
   \FOR{$i = 0, 1, 2, \dotsc$}
   \STATE Roll out trajectories $\tau \sim \pi_{\theta_i}$ 
   \STATE Find $\pi_{\theta_{i+1}}$ minimizing Equation~\eqref{eq:altrposubprobobj}
   \STATE \ \ \ \ \ subject to ${\klbar{\pi_{\theta_{i}}}{\pi_{\theta_{i+1}}} \leq \Delta}$,
   \STATE \ \ \ \ \ with expectations taken over $\tau$ and $\tau_E$ \eqref{eq:avgkl}
   \ENDFOR
\end{algorithmic}
\end{algorithm}


\subsection{Finding cost functions}
\label{sec:cost}

Until now, we deferred discussion of finding cost functions that achieve the supremum in the apprenticeship learning objective~\eqref{eq:alobjective}. We address this issue here for the feature expectation matching setting as described in Section~\ref{sec:featexpmatch}, with cost functions $\mathcal{C}_\text{linear}$ parameterized linearly by $\ell_2$-bounded weight vectors~\eqref{eq:linearcost}. The case for $\mathcal{C}_\text{convex}$ can be derived similarly and is omitted for space reasons.

As mentioned in Section~\ref{sec:policygrad}, the apprenticeship learning algorithm only has access to sample trajectories from $\pi$ and $\pie$, so we will consider finding the cost that achieves the supremum of the empirical apprenticeship learning objective $\hat\delta_\mathcal{C}$~\eqref{eq:empiricaldelta}. Using $\hat c$ to denote the optimal cost for this empirical objective, we have
\begin{align}
\hat\delta_{\mathcal{C}_\text{linear}}(\pi, \pie) 
= \sup_{\|w\|_2 \leq 1} w \cdot (\hat\phi(\pi) - \hat\phi(\pie)),
\end{align}
where $\hat\phi$ is the empirical feature expectation vector. The supremum in this equation is attained by a vector with a closed-form expression:
$
\hat w \triangleq (\hat\phi(\pi) - \hat\phi(\pie))/\| \hat\phi(\pi) - \hat\phi(\pie) \|
$, which can be inserted directly into~\eqref{eq:reinforce} for IM-REINFORCE. However, this $\hat w$ does not suffice for IM-TRPO, which, for the objective of the trust region subproblem~\eqref{eq:altrposubprobobj}, requires a maximizer $\hat w$ that must be recomputed for every optimization step for the subproblem. This recomputation is not difficult or expensive, as we will now demonstrate. For linear costs, the empirical trust region subproblem objective is given by:
\begin{align}
\hat f(\theta) &= \sup_{\| w \|_2 \leq 1} \hat\bbE_{\rho_{\piold}}[w\cdot \phi(s,a)] - \hat\bbE_{\rho_{\pie}}[w \cdot \phi(s,a)]\ + \nonumber \\
&\qquad\hat\bbE_{\rho_{\piold}} \left[ \left(\frac{\pi_\theta(a|s)}{{\piold}(a|s)} - 1\right) Q^w_{\piold}(s,a) \right] \label{eq:lineartrpoprob}
\end{align}
Now let
$
\phi(\piold | s_0,a_0) \triangleq \bbE_{\piold}\left[ \sum_{t=0}^\infty \gamma^t \phi(s_t,a_t) \,\middle|\, s_0,a_0\right]$
and
$
\psiold(\pi_\theta) \triangleq \bbE_{\rho_{\piold}} \left[ \left(\frac{\pi_\theta(a|s)}{{\piold}(a|s)} - 1\right) \phi(\piold|s,a) \right]
$,
both of which are readily estimated from the very same rollout trajectories from $\piold$ used to estimate expected costs. With these, we get $Q^w_{\piold}(s,a) = w\cdot \phi(\piold | s,a)$, which lets us write~\eqref{eq:lineartrpoprob} as:
\begin{align}
\hat f(\theta) &= \sup_{\| w \|_2 \leq 1} w \cdot \left(\hat\phi(\piold) - \hat\phi(\pie)  +  \hat\psiold(\pi_\theta) \right)
\end{align}
This reveals that the supremum is achieved by
\begin{align}
\hat w \triangleq \frac{\hat\phi(\piold) - \hat\phi(\pie)  +  \hat\psiold(\pi_\theta)}{\| \hat\phi(\piold) - \hat\phi(\pie)  +  \hat\psiold(\pi_\theta) \|}, \label{eq:changingw}
\end{align}
which is straightforward to compute. Note that this vector depends on $\theta$; that is, it changes as the trust region subproblem~\eqref{eq:altrpoproblem} is optimized, and must be recomputed with each step of the algorithm for solving the trust region subproblem. However, by construction, all empirical expectations are taken with respect to $\piold$, which does not change as $\theta$ changes, and hence no simulations in the environment are required for these recomputations.


%


\section{Experiments}
\label{sec:experiments}

We evaluated our approach in a variety of scenarios: finite gridworlds of varying sizes, the continuous planar navigation task of Levine and Koltun~\yrcite{levine2012continuous}, a family of continuous environments of varying numbers of observation features~\cite{waterworld}, and a variation of \citeauthor{levine2012continuous}'s highway driving simulation, in which the agent receives high-dimensional egocentric observation features.

In all of the continuous environments, we used policies constructed according to \citet{schulman2015trust}: the policies have Gaussian action distributions, with mean given by a multi-layer perceptron taking observations as input, and standard deviations given by an extra set of parameters. Details on the environments and training methodology are in the supplement.


\paragraph{Comparing against globally optimal methods}
As mentioned in Section~\ref{sec:gametheoretic}, LPAL~\cite{syed2008apprenticeship} finds a global optimum for the apprenticeship problem~\eqref{eq:alprob} with $\mathcal{C}_\text{convex}$ in finite state and action spaces. In contrast, our approach scales to high-dimensional spaces but is only guaranteed to find a local optimum of~\eqref{eq:alprob}, as described in Section~\ref{sec:policyopt}. To evaluate the quality of our local optima, we tested IM-REINFORCE, using $\mathcal{C}_\text{convex}$ to  learn tabular Boltzmann policies with value iteration for exact gradient evaluation, against LPAL and a behavioral cloning baseline.

We evaluated the learned policies for the three algorithms on $64 \times 64$ gridworlds on varying amounts of expert data. In each trial, we randomly generated costs in the world, and we generated expert data by sampling behavior from an optimal policy computed with value iteration. To evaluate an algorithm, we computed the ratio of learned policy performance to the expert's performance. We also ran a timing test, in which we evaluated the computation time for each algorithm on varying gridworld sizes, with fixed dataset sizes, for 10 trials each.

\begin{figure}[ht]
  \vskip -0.1in
  \begin{center}
    \subfigure{\includegraphics[width=0.49\columnwidth]{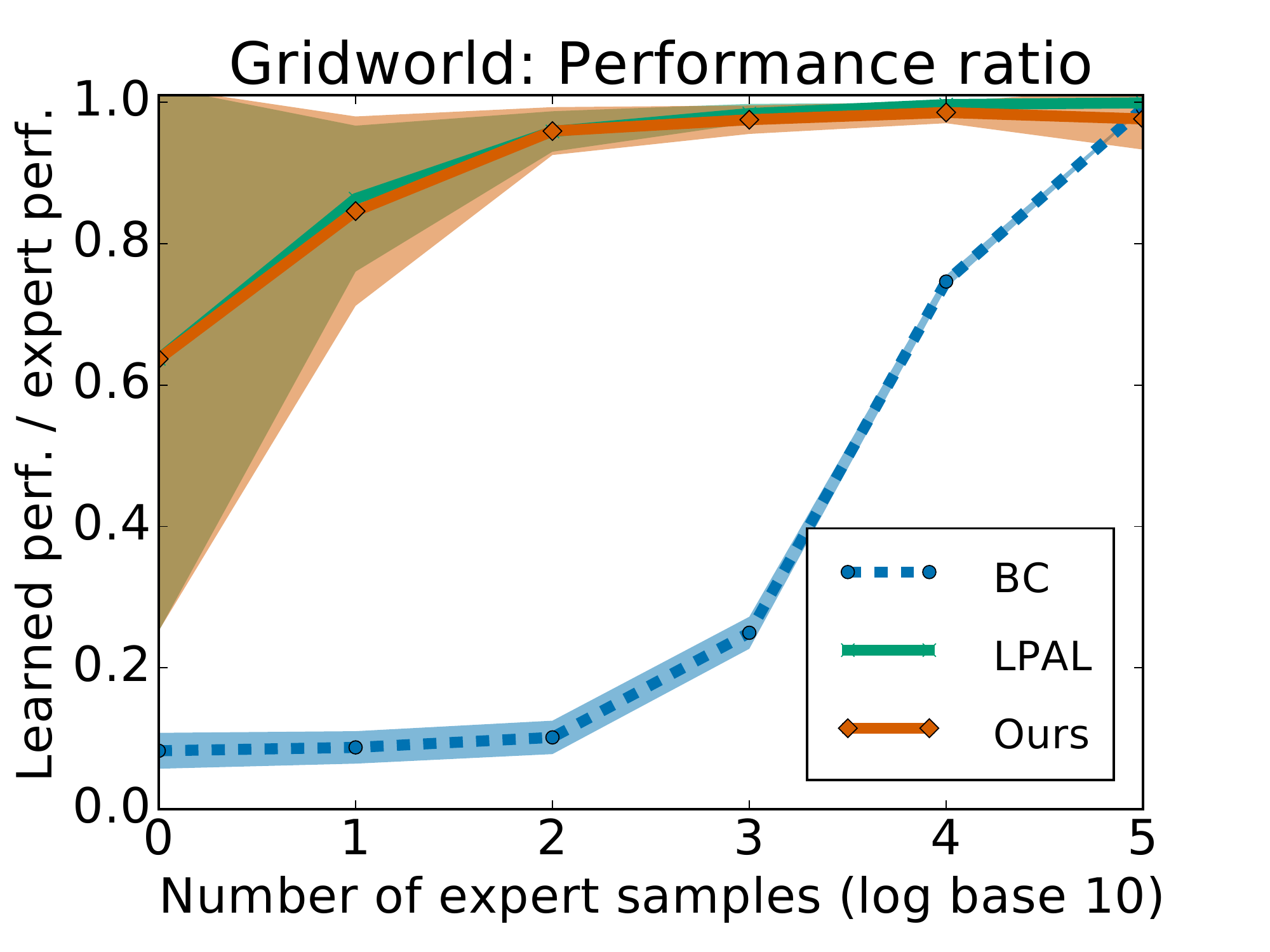}\label{img:gw_perfratio_plot}}
    \subfigure{\includegraphics[width=0.49\columnwidth]{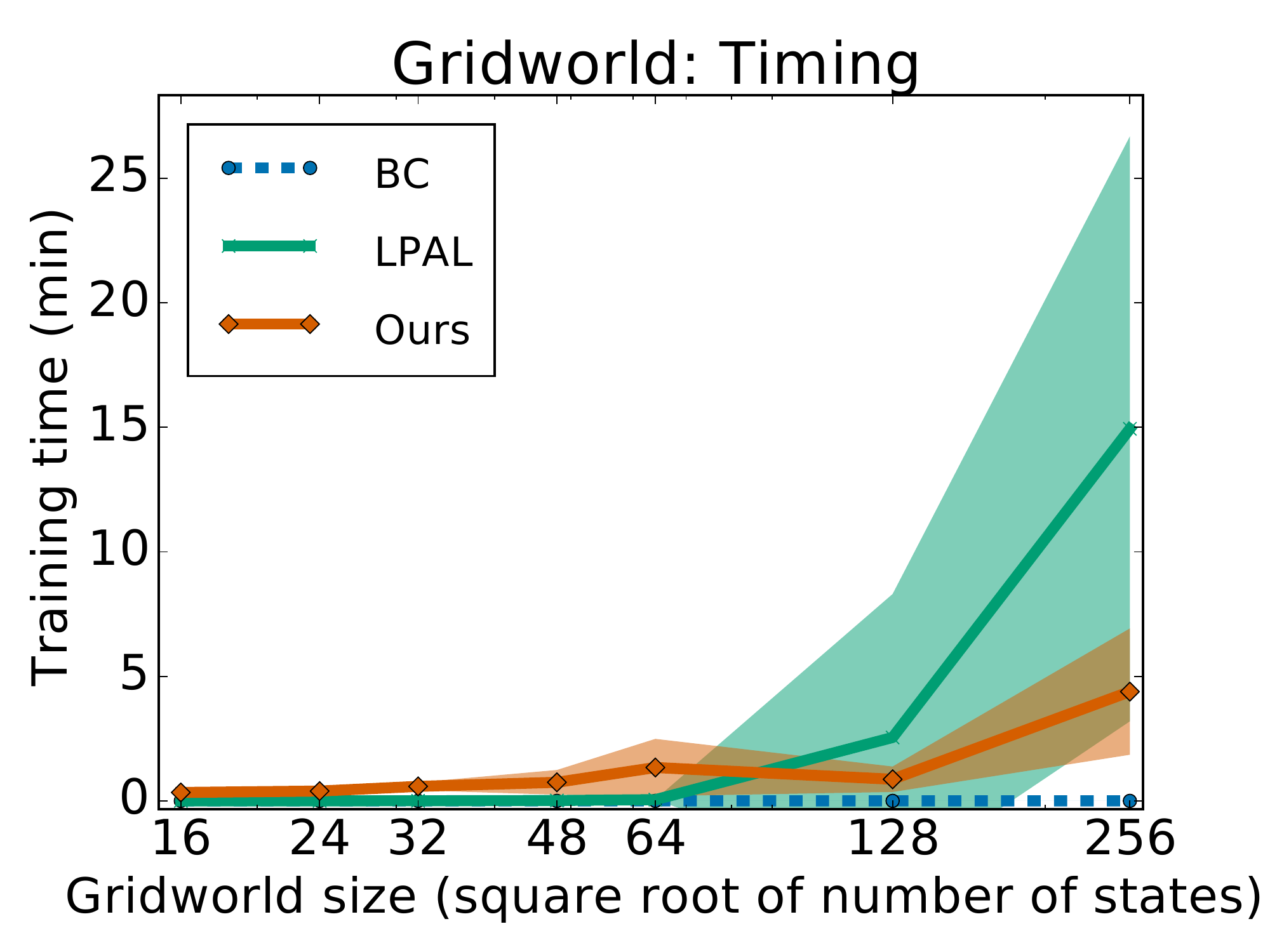}\label{img:gw_timing_plot}}
    \vskip -.1in
    \caption{Left: Gridworld performance ratio across varying amounts of expert data. Right: Training time on increasing gridworld sizes. (\emph{BC} stands for behavioral cloning.)}
    \label{fig:gridworld}
  \end{center}
  \vskip -0.2in
\end{figure}

The results are displayed in Figure~\ref{fig:gridworld}. We found that despite our local optimality guarantee, IM-REINFORCE learned policies achieving at least 98\% the performance of policies learned by LPAL, with similar sample complexity. IM-REINFORCE's training times also scaled favorably compared to LPAL. For a large gridworld with 65536 states, LPAL took on average 10 minutes to train with large variance across instantiations of the expert, whereas our algorithm consistently took around 4 minutes.

\begin{figure}[ht]
   \vskip -0.1in
  \begin{center}
    \subfigure{\includegraphics[width=0.49\columnwidth]{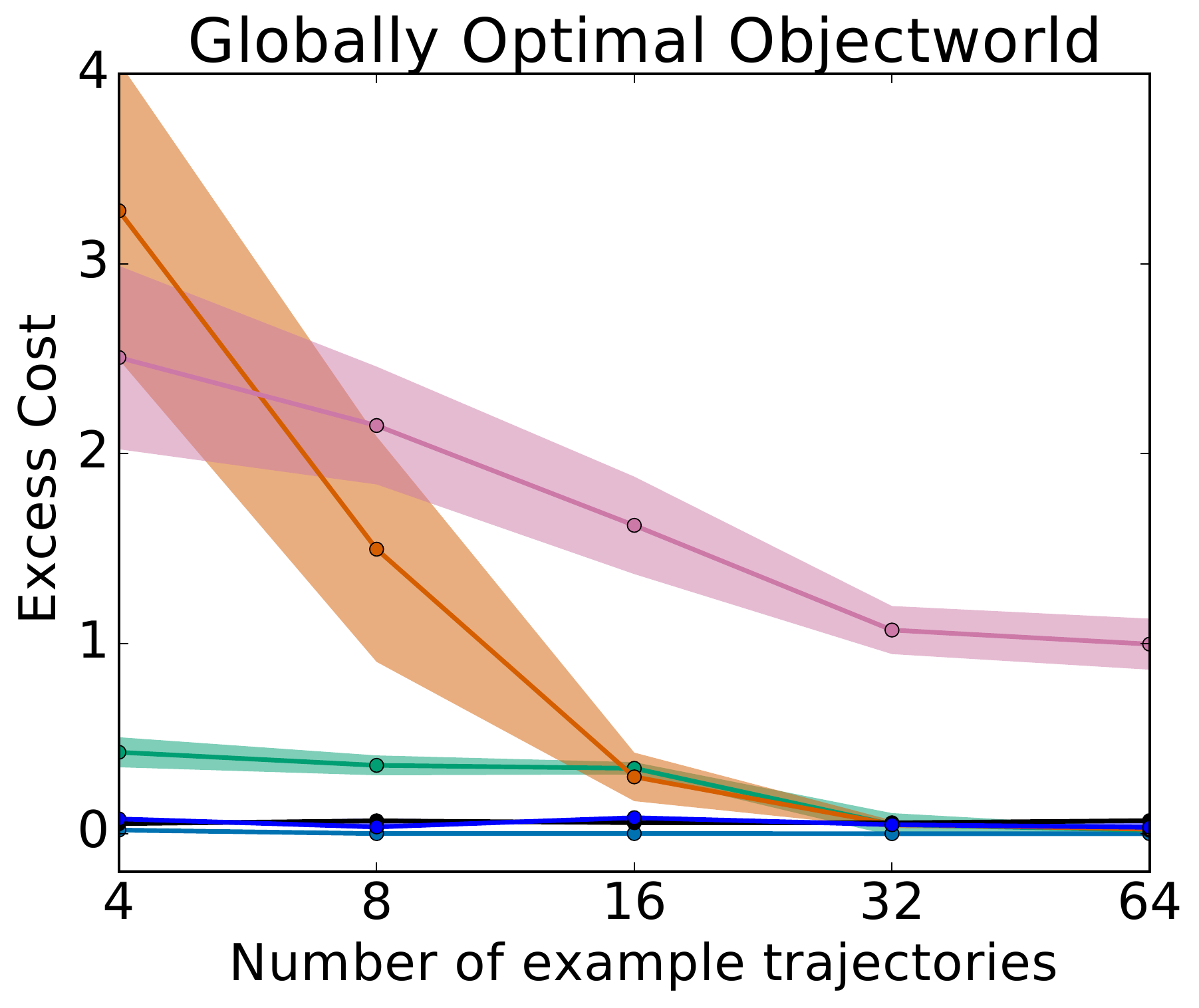}\label{img:rewardloss_global_ow}}
    \subfigure{\includegraphics[width=0.49\columnwidth]{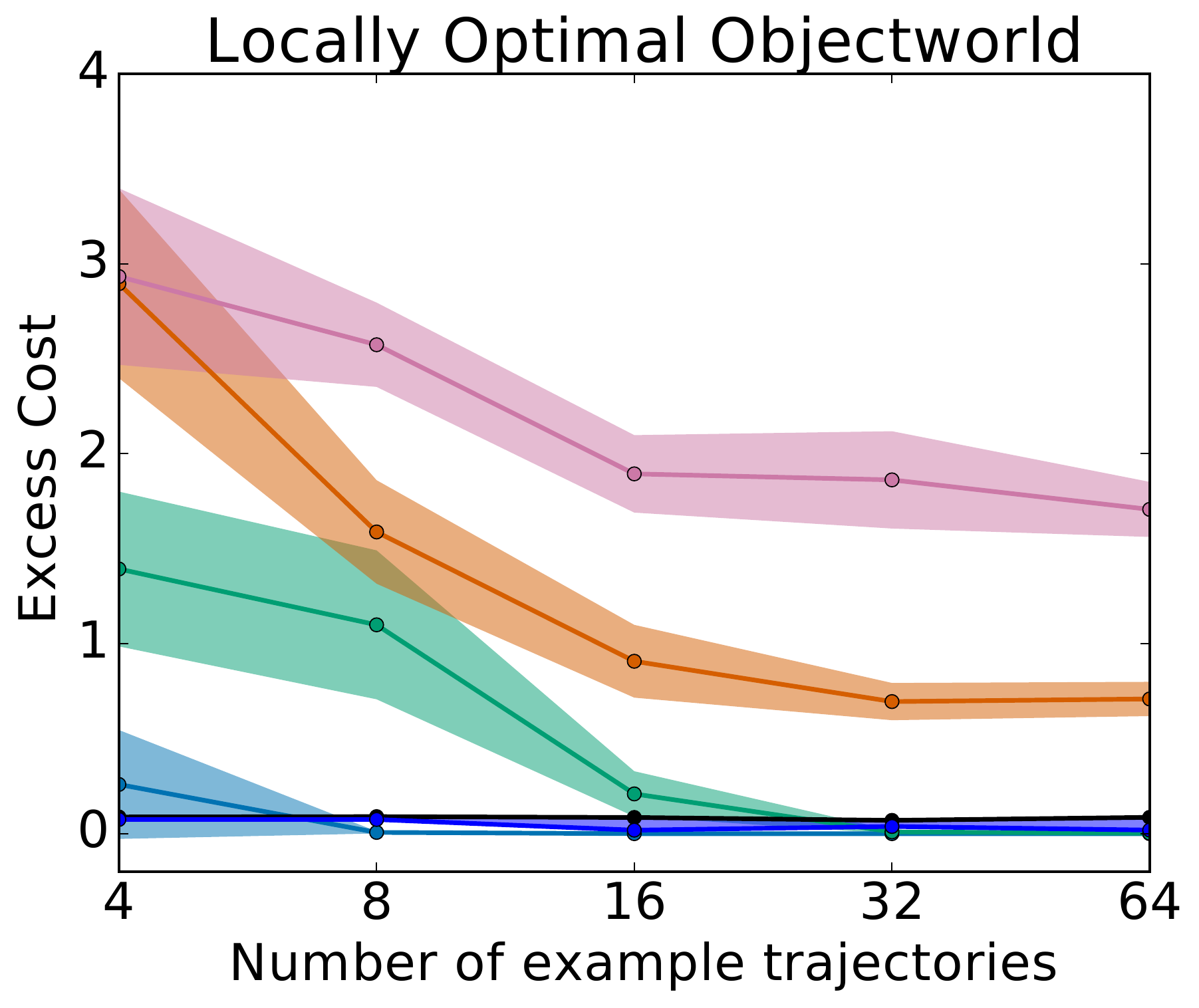}\label{img:rewardloss_local_ow}}
    \subfigure{\includegraphics[width=\columnwidth]{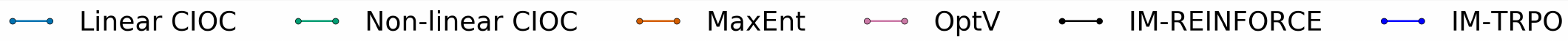}\label{img:obj_legend}}
    \vskip -.1in
    \caption{Excess cost for each algorithm for globally and locally optimal planar navigation examples, against variants of CIOC and other competing algorithms.}
    \label{fig:planar}
  \end{center}
  \vskip -0.2in
\end{figure}

\paragraph{Comparing against continuous IRL}
Next, we evaluated our algorithms in a small, continuous environment: the
objectworld environment of Levine and Koltun~\yrcite{levine2012continuous}, in
which the agent moves in a plane to seek out Gaussian-shaped costs, given only
expert data generated either by globally or locally optimal expert policies. We compared the trajectories produced by IM-TRPO with $\mathcal{C}_\text{linear}$ to those produced by trajectory optimization on a cost learned by Levine and Koltun's CIOC algorithm, a model-based IRL method designed for continuous settings with full knowledge of dynamics derivatives. The basis functions we used for $\mathcal{C}_\text{linear}$ were the same as those used by CIOC to define learned cost functions. The results are in Figure~\ref{fig:planar}.

We found that even though our method is model-free and does not use dynamics derivatives, it consistently learned policies achieving zero excess cost (the difference in expected true cost compared to the expert, measured by averaging over 100 rollouts), matching the performance of optimal trajectories for cost functions learned by CIOC.

\paragraph{Varying dimension}
To evaluate our algorithms' performance with varying environment dimension, we used a family of environments inspired by~\citet{waterworld}. In these environments, the agent moves in a plane populated by colored moving targets to be either captured or avoided, depending on color.
The action space is two-dimensional, allowing the agent to apply forces to move itself in any direction. The agent has a number of sensors $N_\text{sensors}$ facing outward with uniform angular spacing.
Each sensor detects the presence of a target with 5 continuous features indicating the nearest target's distance, color, and relative velocity to the agent.
These sensor features, along with an indicators of whether the agent is currently capturing a target, lead to observation features of dimension $5 \cdot N_\text{sensors} + 2$, which are fed to the policy. Varying $N_\text{sensors}$ yields a family of environments with differing observation dimension.


For $N_\text{sensors}$ set to 5, 10, and 20 (yielding 27, 52, and 102 observation features, respectively), we first generated expert data by executing policies learned by reinforcement learning on a true cost, which was a linear combination of basis functions indicating control effort and intersection with targets. Then, we ran both IM-REINFORCE and IM-TRPO using $\mathcal{C}_\text{linear}$ on the same basis functions, and we measured the excess cost of each learned policy.

\begin{figure}[ht]
\vskip -.15in
\begin{center}
  \subfigure{\includegraphics[width=0.48\columnwidth]{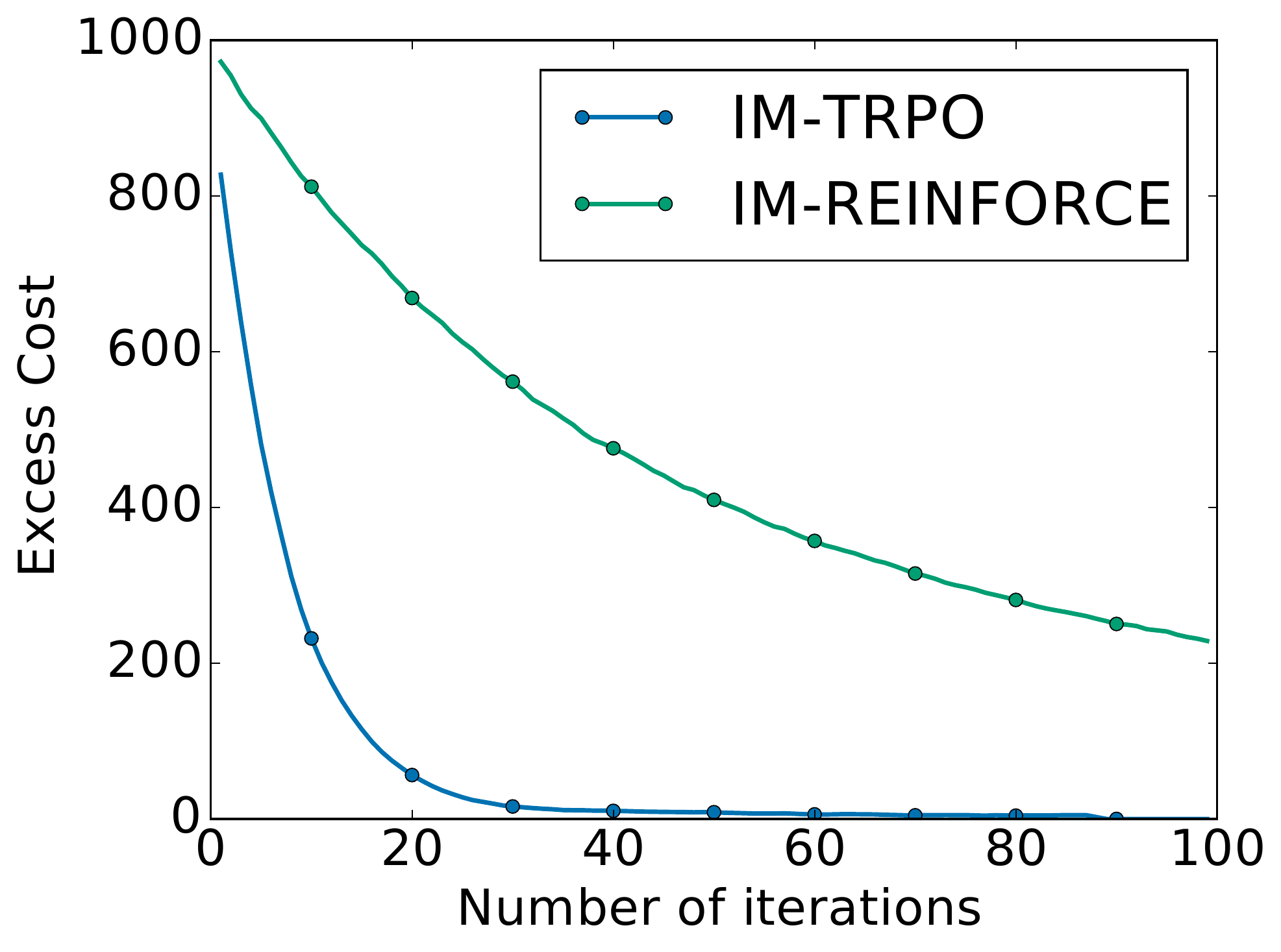}\label{img:rewardloss_water_time}}
  \subfigure{\includegraphics[width=0.48\columnwidth]{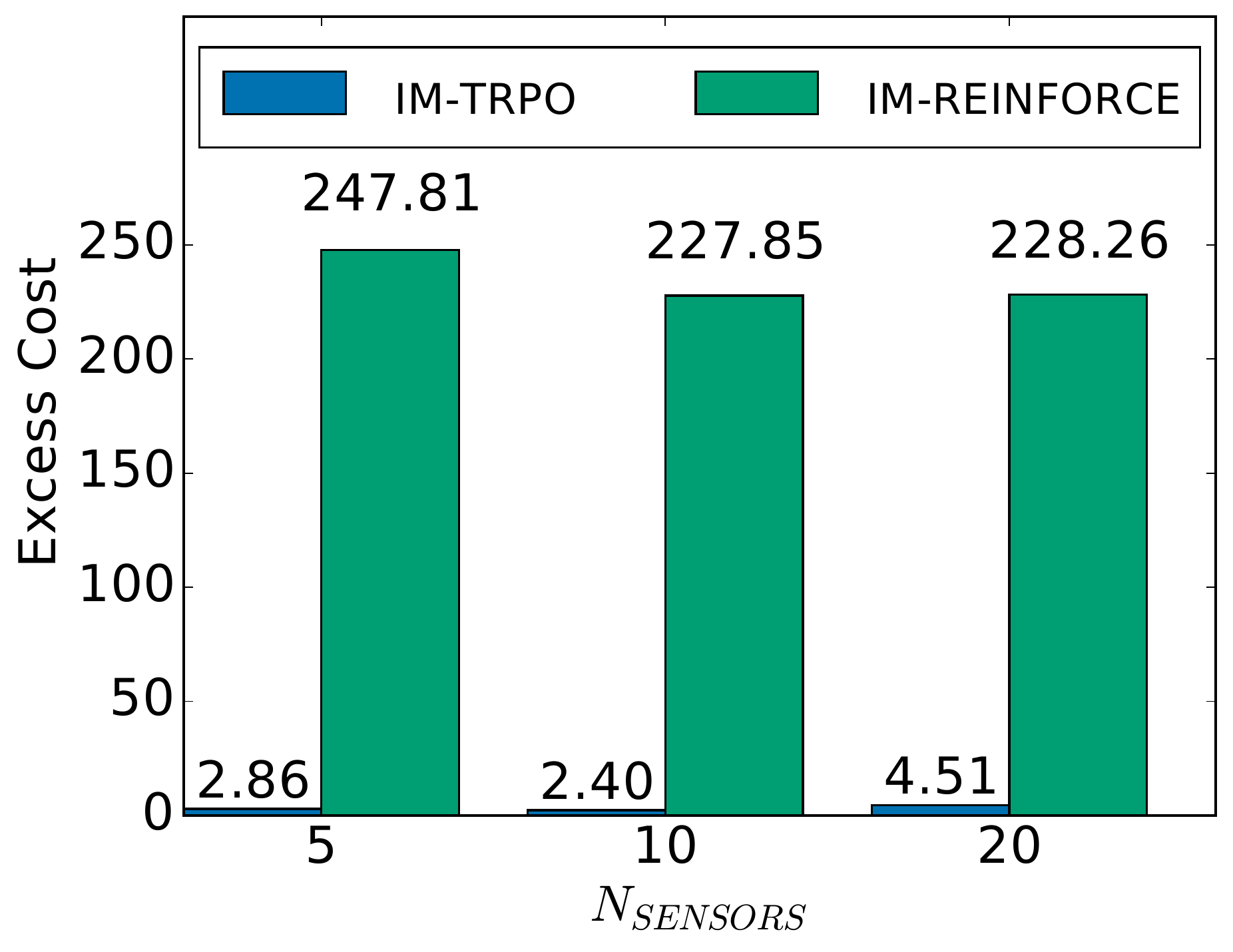}\label{img:rewardloss_water}}
    \vskip -.1in
  \caption{Left: Excess cost over time for one run of
    ${N_\text{sensors} = 20}$. Curves for other settings are similar. Right: Excess costs for learned policies on various sensor counts.} \label{fig:waterworld}
\end{center}
\vskip -0.1in
\end{figure}

We found that IM-TRPO achieved nearly perfect imitation in this setting, and the performance was not significantly affected by the dimensionality of the space. IM-REINFORCE's learning also progressed, but was far outpaced by IM-TRPO (see Figure~\ref{fig:waterworld}). 
We also verified that IM-TRPO's overhead of computing $\hat c$~\eqref{eq:changingw} for each step of solving its trust region subproblem was negligible, verifying our claim in Section~\ref{sec:cost}. On our system, iterations for plain TRPO for reinforcement learning and IM-TRPO both took 8-9 seconds each for this environment, with no statistically significant difference.

\begin{figure}[ht]
\vskip -0.05in
\begin{center}
  \includegraphics[width=.8\columnwidth]{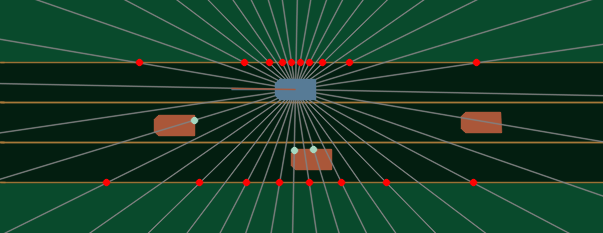}
  \vskip -.1in
  \caption{Car sensors for highway driving. The blue car is the agent, and emanating lines indicate sensor directions. Some sensors see lanes (red points), and others see neighboring cars (cyan points).} \label{fig:highwaysensors}
\end{center}
\vskip -0.025in
\end{figure}

\paragraph{Highway driving}
Finally, we ran IM-TRPO on a variation of the highway driving task
of~\citet{levine2012continuous}. In this task, the learner must imitate driving
behaviors (aggressive, tailgating, and evasive) in a continuous driving
simulation. The observations in the original driving task were the actual states
of the whole environment, including positions and velocities of all cars at all
points on the road. To introduce more realism, we modified the environment by
providing policies only egocentric observation features: readings from 30
equally spaced rangefinders that detect the two road edges, readings from 60
equally spaced rangefinders that detect cars, and speed and angular velocity of the agent's car. These observations effectively form a depth image of nearby cars and lane markings within the agent's field of view; see Figure~\ref{fig:highwaysensors}. We aggregated these readings over a window of 5 timesteps, yielding a 610-dimensional partial observations.

We ran IM-TRPO with $\mathcal{C}_\text{linear}$, using basis functions representing quadratic features derived from those of~\citeauthor{levine2012continuous}. Despite the fact that we provided only high-dimensional partial observations, our model-free approach learned policies achieving behavior comparable to the trajectories generated by CIOC, which was provided full state features and a full environment model. The policies learned by our algorithm generated behavior that both qualitatively and quantitatively resembled the demonstrated behavior, as shown in Table~\ref{table:highway}.

\begin{table}[ht]
\vskip -10pt
\caption{Statistics for sample trajectories for IM-TRPO, compared to CIOC~\cite{levine2012continuous} and human demonstrations. IM-TRPO and CIOC both generate human-like behavior.} \label{table:highway}
\begin{center}
\scriptsize
\begin{sc}
  \vskip -10pt
\begin{tabular}{lcccc}
\hline
\abovespace
\belowspace
Style & Path & \begin{tabular}{@{}c@{}}Avg. speed \\ (km/h)\end{tabular} & \begin{tabular}{@{}c@{}}Time \\ behind (s)\end{tabular} & \begin{tabular}{@{}c@{}}Time \\ in front (s)\end{tabular}\\
\hline
\abovespace
Aggressive & Human   & 158.2 & 3.5   & 16.7 \\
 & CIOC   & 158.1 &  3.5   & 12.5 \\
\belowspace
 & IM-TRPO & 147.8 & 4.2 & 9.2 \\
Evasive & Human   & 149.5 & 4.5   & 2.8 \\
 & CIOC   & 150.1 &  7.2   & 3.7 \\
\belowspace
 & IM-TRPO & 110.4 & 4.6 & 3.9 \\
Tailgater & Human   & 115.3 & 99.5   & 7.0 \\
 & CIOC   & 97.5 &  111.0   & 0.0 \\
\belowspace
 & IM-TRPO & 97.6 & 71.4 & 12.3 \\
\hline
\end{tabular}
\end{sc}
\end{center}
 \vskip -0.2in
\end{table}

\section{Discussion and future work}
We showed that carefully blending state-of-the-art policy gradient algorithms for reinforcement learning with local cost function fitting lets us successfully train neural network policies for imitation in high-dimensional, continuous environments. Our method is able to identify a locally optimal solution, even in settings where optimal planning is out of reach. This is a significant advantage over  competing algorithms that require repeatedly solving planning problems in an inner loop. In fact, when the inner planning problem is only approximately solved, competing algorithms do not even provide local optimality guarantees~\cite{ermon2015learning}.

Our approach does not use expert interaction or reinforcement signal, fitting in a family of such approaches that includes apprenticeship learning and inverse reinforcement learning. When either of these additional resources is provided, alternative approaches~\cite{kim2013learning,daume2009search,ross2010efficient,Ross2010-wn} may be more sample efficient, and investigating ways to combine these resources with our framework is an interesting research direction.

We focused on the policy optimization component of apprenticeship learning, rather than the design of appropriate cost function classes. 
We believe this is an important area for future work. Nonlinear cost function classes have been successful in IRL~\cite{ratliff2009learning,levine2011nonlinear} as well as in other machine learning problems reminiscent of ours, in particular that of training generative image models. In the language of generative adversarial networks~\cite{goodfellow2014generative}, the policy parameterizes a generative model of state-action pairs, and the cost function serves as an adversary. Apprenticeship learning with large cost function classes capable of distinguishing between arbitrary state-action visitation distributions would, enticingly, open up the possibility of exact imitation.



\section*{Acknowledgements}
We thank John Schulman for valuable conversations about TRPO. This work was supported by a grant from the SAIL-Toyota Center for AI Research and by a National Science Foundation Graduate Research Fellowship (grant no. DGE-114747).

\bibliography{paper}

\begin{thebibliography}{26}
\providecommand{\natexlab}[1]{#1}
\providecommand{\url}[1]{\texttt{#1}}
\expandafter\ifx\csname urlstyle\endcsname\relax
  \providecommand{\doi}[1]{doi: #1}\else
  \providecommand{\doi}{doi: \begingroup \urlstyle{rm}\Url}\fi

\bibitem[Abbeel \& Ng(2004)Abbeel and Ng]{Abbeel2004-cf}
Abbeel, Pieter and Ng, Andrew~Y.
\newblock Apprenticeship learning via inverse reinforcement learning.
\newblock In \emph{Proceedings of the 21st International Conference on Machine
  Learning}, 2004.

\bibitem[Bagnell(2015)]{bagnell2015invitation}
Bagnell, J~Andrew.
\newblock An invitation to imitation.
\newblock Technical report, Carnegie Mellon University, 2015.

\bibitem[Daum{\'e}~{III} et~al.(2009)Daum{\'e}~{III}, Langford, and
  Marcu]{daume2009search}
Daum{\'e}~{III}, Hal, Langford, John, and Marcu, Daniel.
\newblock Search-based structured prediction.
\newblock \emph{Machine learning}, 75\penalty0 (3):\penalty0 297--325, 2009.

\bibitem[Ermon et~al.(2015)Ermon, Xue, Toth, Dilkina, Bernstein, Damoulas,
  Clark, DeGloria, Mude, Barrett, et~al.]{ermon2015learning}
Ermon, Stefano, Xue, Yexiang, Toth, Russell, Dilkina, Bistra~N, Bernstein,
  Richard, Damoulas, Theodoros, Clark, Patrick, DeGloria, Steve, Mude, Andrew,
  Barrett, Christopher, et~al.
\newblock Learning large-scale dynamic discrete choice models of
  spatio-temporal preferences with application to migratory pastoralism in
  {E}ast {A}frica.
\newblock In \emph{AAAI}, pp.\  644--650, 2015.

\bibitem[Goodfellow et~al.(2014)Goodfellow, Pouget-Abadie, Mirza, Xu,
  Warde-Farley, Ozair, Courville, and Bengio]{goodfellow2014generative}
Goodfellow, Ian, Pouget-Abadie, Jean, Mirza, Mehdi, Xu, Bing, Warde-Farley,
  David, Ozair, Sherjil, Courville, Aaron, and Bengio, Yoshua.
\newblock Generative adversarial nets.
\newblock In \emph{Advances in Neural Information Processing Systems}, pp.\
  2672--2680, 2014.

\bibitem[Kakade \& Langford(2002)Kakade and Langford]{kakade2002approximately}
Kakade, Sham and Langford, John.
\newblock Approximately optimal approximate reinforcement learning.
\newblock In \emph{Proceedings of the 19th International Conference on Machine
  Learning}, pp.\  267--274, 2002.

\bibitem[Karpathy(2015)]{waterworld}
Karpathy, Andrej.
\newblock Reinforcejs: Waterworld demo, 2015.
\newblock URL
  \url{http://cs.stanford.edu/people/karpathy/reinforcejs/waterworld.html}.

\bibitem[Kim et~al.(2013)Kim, Farahmand, Pineau, and Precup]{kim2013learning}
Kim, Beomjoon, Farahmand, Amir-massoud, Pineau, Joelle, and Precup, Doina.
\newblock Learning from limited demonstrations.
\newblock In \emph{Advances in Neural Information Processing Systems}, pp.\
  2859--2867, 2013.

\bibitem[Levine \& Koltun(2012)Levine and Koltun]{levine2012continuous}
Levine, Sergey and Koltun, Vladlen.
\newblock Continuous inverse optimal control with locally optimal examples.
\newblock In \emph{Proceedings of the 29th International Conference on Machine
  Learning}, pp.\  41--48, 2012.

\bibitem[Levine et~al.(2011)Levine, Popovic, and Koltun]{levine2011nonlinear}
Levine, Sergey, Popovic, Zoran, and Koltun, Vladlen.
\newblock Nonlinear inverse reinforcement learning with gaussian processes.
\newblock In \emph{Advances in Neural Information Processing Systems}, pp.\
  19--27, 2011.

\bibitem[Neu \& Szepesv\'{a}ri(2009)Neu and Szepesv\'{a}ri]{Neu2009-xe}
Neu, Gergely and Szepesv\'{a}ri, Csaba.
\newblock Training parsers by inverse reinforcement learning.
\newblock \emph{Mach. Learn.}, 77\penalty0 (2-3):\penalty0 303--337, 11~April
  2009.

\bibitem[Ng \& Russell(2000)Ng and Russell]{ng2000algorithms}
Ng, Andrew~Y and Russell, Stuart~J.
\newblock Algorithms for inverse reinforcement learning.
\newblock In \emph{Proceedings of the 17th International Conference on Machine
  Learning}, pp.\  663--670, 2000.

\bibitem[Peters \& Schaal(2008)Peters and Schaal]{peters2008reinforcement}
Peters, Jan and Schaal, Stefan.
\newblock Reinforcement learning of motor skills with policy gradients.
\newblock \emph{Neural networks}, 21\penalty0 (4):\penalty0 682--697, 2008.

\bibitem[Pomerleau(1991)]{pomerleau1991efficient}
Pomerleau, Dean~A.
\newblock Efficient training of artificial neural networks for autonomous
  navigation.
\newblock \emph{Neural Computation}, 3\penalty0 (1):\penalty0 88--97, 1991.

\bibitem[Ratliff et~al.(2006)Ratliff, Bagnell, and Zinkevich]{Ratliff2006-bm}
Ratliff, Nathan~D, Bagnell, J~Andrew, and Zinkevich, Martin~A.
\newblock Maximum margin planning.
\newblock In \emph{Proceedings of the 23rd International Conference on Machine
  Learning}, pp.\  729--736, 2006.

\bibitem[Ratliff et~al.(2009)Ratliff, Silver, and Bagnell]{ratliff2009learning}
Ratliff, Nathan~D, Silver, David, and Bagnell, J~Andrew.
\newblock Learning to search: Functional gradient techniques for imitation
  learning.
\newblock \emph{Autonomous Robots}, 27\penalty0 (1):\penalty0 25--53, 2009.

\bibitem[Ross \& Bagnell(2010)Ross and Bagnell]{ross2010efficient}
Ross, St{\'e}phane and Bagnell, Drew.
\newblock Efficient reductions for imitation learning.
\newblock In \emph{International Conference on Artificial Intelligence and
  Statistics}, pp.\  661--668, 2010.

\bibitem[Ross et~al.(2011)Ross, Gordon, and Bagnell]{Ross2010-wn}
Ross, St{\'e}phane, Gordon, Geoffrey~J, and Bagnell, Drew.
\newblock A reduction of imitation learning and structured prediction to
  no-regret online learning.
\newblock In \emph{International Conference on Artificial Intelligence and
  Statistics}, pp.\  627--635, 2011.

\bibitem[Russell(1998)]{russell1998learning}
Russell, Stuart.
\newblock Learning agents for uncertain environments.
\newblock In \emph{Proceedings of the Eleventh Annual Conference on
  Computational Learning Theory}, pp.\  101--103, 1998.

\bibitem[Schulman et~al.(2015)Schulman, Levine, Abbeel, Jordan, and
  Moritz]{schulman2015trust}
Schulman, John, Levine, Sergey, Abbeel, Pieter, Jordan, Michael, and Moritz,
  Philipp.
\newblock Trust region policy optimization.
\newblock In \emph{Proceedings of the 32nd International Conference on Machine
  Learning}, pp.\  1889--1897, 2015.

\bibitem[Sutton et~al.(1999)Sutton, McAllester, Singh, and
  Mansour]{sutton1999policy}
Sutton, Richard~S, McAllester, David~A, Singh, Satinder~P, and Mansour, Yishay.
\newblock Policy gradient methods for reinforcement learning with function
  approximation.
\newblock In \emph{Advances in Neural Information Processing Systems}, pp.\
  1057--1063, 1999.

\bibitem[Syed \& Schapire(2007)Syed and Schapire]{syed2007game}
Syed, Umar and Schapire, Robert~E.
\newblock A game-theoretic approach to apprenticeship learning.
\newblock In \emph{Advances in Neural Information Processing Systems}, pp.\
  1449--1456, 2007.

\bibitem[Syed \& Schapire(2010)Syed and Schapire]{syed2010reduction}
Syed, Umar and Schapire, Robert~E.
\newblock A reduction from apprenticeship learning to classification.
\newblock In \emph{Advances in Neural Information Processing Systems}, pp.\
  2253--2261, 2010.

\bibitem[Syed et~al.(2008)Syed, Bowling, and Schapire]{syed2008apprenticeship}
Syed, Umar, Bowling, Michael, and Schapire, Robert~E.
\newblock Apprenticeship learning using linear programming.
\newblock In \emph{Proceedings of the 25th International Conference on Machine
  Learning}, pp.\  1032--1039, 2008.

\bibitem[Williams(1992)]{williams1992simple}
Williams, Ronald~J.
\newblock Simple statistical gradient-following algorithms for connectionist
  reinforcement learning.
\newblock \emph{Machine learning}, 8\penalty0 (3-4):\penalty0 229--256, 1992.

\bibitem[Ziebart et~al.(2008)Ziebart, Maas, Bagnell, and
  Dey]{ziebart2008maximum}
Ziebart, Brian~D, Maas, Andrew~L, Bagnell, J~Andrew, and Dey, Anind~K.
\newblock Maximum entropy inverse reinforcement learning.
\newblock In \emph{AAAI}, pp.\  1433--1438, 2008.

\end{thebibliography}
\bibliographystyle{icml2016}

\newpage
\twocolumn[
\icmltitle{\mytitle: \\Supplementary Material}

\icmlauthor{Jonathan Ho}{hoj@cs.stanford.edu}
\icmlauthor{Jayesh K. Gupta}{jkg@cs.stanford.edu}
\icmlauthor{Stefano Ermon}{ermon@cs.stanford.edu}
\icmladdress{Stanford University}

\icmlkeywords{imitation learning, apprenticeship learning, inverse reinforcement learning, inverse optimal control}

\vskip 0.3in

Here, we give extra information regarding the environment and algorithm setups for our experiments.

\paragraph{Gridworld}
We used tabular policies for IM-REINFORCE with parameters $\theta_{sa}$, with action probabilities $\pi(a|s) \propto \exp(\theta_{sa})$; we used value iteration to obtain $Q$ values for the gradient formula~\eqref{eq:reinforce}. We solved the linear programs for LPAL with Gurobi 6.5.1, and we defined the policies learned by behavioral cloning as simple lookups into expert data (for states unseen in the expert data, a random action is chosen). All timing tests were performed on an 4-core 3.6GHz Intel i7-4790 CPU.

The gridworlds we used resembled those of Abbeel and Ng~\yrcite{Abbeel2004-cf}. Each was a square grid of states, with five actions (an action to move in each compass direction, and one for staying in place) that fail with 30\% probability and result in a random move. Each test consisted of 40 trials. Costs were generated in $8 \times 8$ non-overlapping regions in the gridworld, giving one basis function for $\mathcal{C}_\text{convex}$ per region.

\paragraph{Waterworld}
We first ran TRPO for various iteration counts to obtain expert policies achieving various expected costs according to the true cost function, which penalized application of control, and assigned differing cost values to the targets of different colors. Then, we executed each expert policy to yield 25 trajectory samples, and then we ran IM-REINFORCE and IM-TRPO both for 100 iterations to imitate each expert policy. The trajectories were 500 timesteps long, and the discount factor was 0.99. We gave both algorithms 50 rollouts per iteration. Excess costs were computed by averaging over 100 rollouts.

\paragraph{Highway}
For each driving style, we ran IM-TRPO for 500 iterations, each collecting 20000 state-action pairs with simulation. The datasets and dynamics model were identical to the ones used by~\citet{levine2012continuous}. We evaluated our policies with the same measurements used by \citeauthor{levine2012continuous}, averaged over 50 rollouts.
]

\end{document}